\documentclass[sigconf]{acmart}
\AtBeginDocument{%
  \providecommand\BibTeX{{%
    \normalfont B\kern-0.5em{\scshape i\kern-0.25em b}\kern-0.8em\TeX}}}

\acmSubmissionID{1414}

\usepackage{graphicx}

\usepackage{enumitem}
\usepackage{balance}

\copyrightyear{2024}
\acmYear{2024}
\setcopyright{rightsretained}
\acmConference[HRI '24]{Proceedings of the 2024 ACM/IEEE International Conference on Human-Robot Interaction}{March 11--14, 2024}{Boulder, CO, USA} \acmBooktitle{Proceedings of the 2024 ACM/IEEE International Conference on Human-Robot Interaction (HRI '24), March 11--14, 2024, Boulder, CO, USA} \acmDOI{10.1145/3610977.3634965} \acmISBN{979-8-4007-0322-5/24/03}

\makeatletter
\gdef\@copyrightpermission{
  \begin{minipage}{0.3\columnwidth}
   \href{https://creativecommons.org/licenses/by/4.0/}{\includegraphics[width=0.90\textwidth]{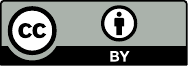}}
  \end{minipage}\hfill
  \begin{minipage}{0.7\columnwidth}
   \href{https://creativecommons.org/licenses/by/4.0/}{This work is licensed under a Creative Commons Attribution International 4.0 License.}
  \end{minipage}
  \vspace{5pt}
}
\makeatother

\begin{document}

\title{A System for Human-Robot Teaming through End-User Programming and Shared Autonomy}

\author{Michael Hagenow}
\orcid{1234-5678-9012}
\affiliation{%
  \institution{Massachusetts Institute of Technology}
  \streetaddress{70 Vassar St}
  \city{Cambridge}
  \state{Massachusetts}
  \country{USA}
  \postcode{02476}
}
\email{hagenow@mit.edu}

\author{Emmanuel Senft}
\affiliation{%
  \institution{Idiap Research Institute}
  \city{Martigny}
  \country{Switzerland}}
\email{esenft@idiap.ch}

\author{Robert Radwin}
\affiliation{%
  \institution{University of Wisconsin -- Madison}
  \city{Madison}
  \state{Wisconsin}
  \country{USA}}
\email{rradwin@wisc.edu}

\author{Michael Gleicher}
\affiliation{%
  \institution{University of Wisconsin -- Madison}
  \city{Madison}
  \state{Wisconsin}
  \country{USA}}
\email{gleicher@cs.wisc.edu}

\author{Michael Zinn}
\affiliation{%
  \institution{University of Wisconsin -- Madison}
  \city{Madison}
  \state{Wisconsin}
  \country{USA}}
\email{mike.zinn@wisc.edu}

\author{Bilge Mutlu}
\affiliation{%
  \institution{University of Wisconsin -- Madison}
  \city{Madison}
  \state{Wisconsin}
  \country{USA}}
\email{bilge@cs.wisc.edu}

\renewcommand{\shortauthors}{Michael Hagenow et al.}

\begin{abstract}
     Many industrial tasks---such as sanding, installing fasteners, and wire harnessing---are difficult to automate due to task complexity and variability. We instead investigate deploying robots in an assistive role for these tasks, where the robot assumes the physical task burden and the skilled worker provides both the high-level task planning and low-level feedback necessary to effectively complete the task. In this article, we describe the development of a system for flexible human-robot teaming that combines state-of-the-art methods in end-user programming and shared autonomy and its implementation in sanding applications. We demonstrate the use of the system in two types of sanding tasks, situated in aircraft manufacturing, that highlight two potential workflows within the human-robot teaming setup. We conclude by discussing challenges and opportunities in human-robot teaming identified during the development, application, and demonstration of our system.
\end{abstract}





\maketitle

\section{Introduction}

\begin{figure}[]
\centering
\includegraphics[width=0.48\textwidth]{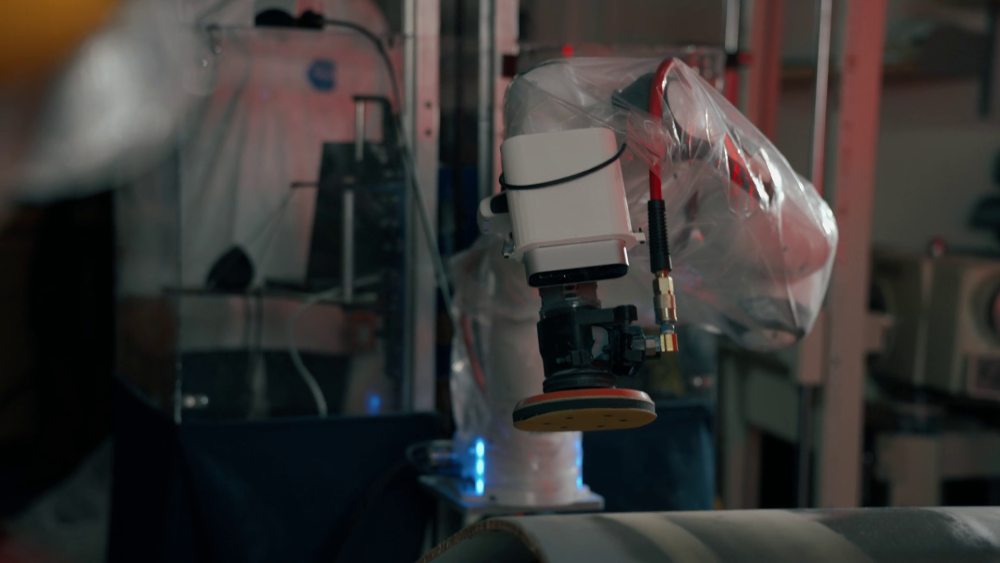}
\Description{A picture of the collaborative sanding robot where you can see the orbital sanding end effector and camera. The operator is in a white coat in the background.}
\caption{In this paper, we describe the development of a human-robot teaming solution for variable industrial sanding tasks that combines the strengths of the operator and robot through end-user programming and shared autonomy.}
\label{fig:teaser}
\vspace{-15pt}
\end{figure}

Collaborative robots, or \emph{cobots}, are designed to work in human environments and offer new ways to engage with industrial workers. For example, a cobot can serve as an intelligent teammate, offloading physically demanding or repetitive actions across a range of tasks. However, the use of cobots in practice falls short of this vision. With currently available methods to task robots (e.g., teach pendants), the majority of collaborative robots are programmed to execute basic skills, such as pick and place, and deployed to primitive tasks such as machine tending \cite{michaelis2020collaborative}. Looking beyond these basic applications, there are opportunities to deploy cobots to more complex and physically demanding jobs, such as sanding or sealant application, to minimize hazardous worker conditions (e.g., high force loads, enclosed spaces). Moving toward complex industrial tasks requires moving from highly structured and deterministic task settings (e.g., moving objects in known locations) to less structured and highly variable task settings (e.g., sanding a surface with intermittent defects). This shift poses a number of challenges for task automation. The required robustness and planning for variability is outside the scope of traditional task automation pipelines and current research technologies are not yet enabling reliable automation for many complex, contact-rich tasks \cite{suomalainen2022survey}. For example, in auto body repair, each sanding task will be a different shape and require removing different amounts of damaged material. As human workers already possess the expertise (e.g., sensing, reasoning) required to complete such tasks, there is a need for \emph{human-in-the-loop} approaches, where a skilled operator controls the robot to assure reliable task outcomes. In this paper, we propose a \textit{human-robot teaming} system that balances the task workload to leverage the strengths of the robot, such as precision and offloading the physical task burden, and the strengths of the human operator, including reasoning about the task parameters and planning.

The split of workload in human-robot teams is often identified by the level of automation \cite{parasuraman2000model}, which provides a score for how often and through what means a human interacts with a robot system. Examples of these levels include manual teleoperation, shared control, supervisory control, and full automation. Within these levels, the human and robot assume different roles; such as \emph{acting}, \emph{deciding}, and \emph{suggesting} \cite{endsley1995out}. Many methods select fixed human and robot roles based on the needs of a particular task. In limited cases, methods allow role-shifting throughout a task based on task and user modeling \cite{abbink2012haptic,milliken2017modeling} to address different requirements for human intervention. Following a similar premise, we posit that successfully addressing variability in complex industrial tasks requires skilled human teammates to interact with the system at multiple times with varying levels of feedback. Rather than pursuing techniques for role shifting, we base our human-robot teaming approach on explicitly identified needs for the skilled worker throughout industrial tasks. Specifically, we desire to leverage the reasoning capabilities of the human to help define the task and the low-level action planning and perception of the human to react to variability when the process outcome is unexpected. Our resulting system meets these needs by fusing two modern techniques within the human-machine interface: end-user programming \cite{ajaykumar2021survey} and shared autonomy \cite{selvaggio2021autonomy}.

In contrast to previous work, we target separate types of human interaction for the \emph{specification} and \emph{execution} phases of a task. End-user programming aims to develop intuitive interfaces for task specification that remove the burden of traditional robot programming while maintaining expressiveness for operators to specify nuanced and complex task plans. Shared autonomy aims to tightly couple the human in the robot's control loop such that the human and robot can work together to plan appropriate actions during task execution. By combining techniques that address the different elements of task variability, the resulting system can allow operators to provide sufficient feedback to ensure reliable task outcomes.

This paper describes our development of a system for complex industrial tasks that leverages the respective strengths of a human operator and a robot. The main contributions of this work are:

\begin{itemize}[leftmargin=*]
    \item Describing \textit{a novel human-robot teaming approach} that combines techniques in end-user programming and shared autonomy to achieve flexible automation for variable tasks;
    \item Developing \textit{a prototype system and corresponding workflows} that instantiates the human-robot teaming approach to complete representative sanding tasks in aviation manufacturing. 
\end{itemize}


We believe that systems enabling flexible automation can extend the use cases of cobots (including ergonomically hazardous tasks like sanding) and increase adoption of cobot technology. In the remaining sections, we first describe motivating scenarios in aviation manufacturing and resulting technological requirements for our system. We then describe our prototype implementation and system workflows, which were used to complete representative tasks in both a lab setting and on-site at an aviation manufacturing facility. We conclude by discussing results and opportunities.
\section{Related Work}

\subsection{Industrial Human-Robot Collaboration}
Given the relative advantages of robots over human workers, such as precision and repeatability, industrial robots have become pervasive in modern manufacturing. Industrial robots are deployed across a range of jobs, including handling, welding, assembling, and painting \citep{hagele2016industrial}. While in most cases, robots operate autonomously in sequestered spaces and often employ custom hardware and tooling (e.g., gantry systems for automated tape layup in aviation \citep{sarh2009aircraft}), there is a desire to develop methods that promote flexibility and collaboration through the use of  general-purpose robotic hardware (e.g., cobots). For example, Fujii et al. \cite{fujii2016study} developed a semi-autonomous system where an operator hand guides an industrial robot when handling large components and the robot also completes some automatic subtasks, like fetching materials. Carmichael et al. \cite{carmichael2019anbot} similarly created a system for abrasive blasting where an operator drives the robot behavior via physical interaction while the robot assumes the physical burden of the task. Maric et al. \cite{maric2020collaborative} apply a similar paradigm where the human guides a robot through sanding of complex surfaces, from which the robot learns object-centric trajectories. Other methods focus on scheduling of interdependent tasks and timing in human-robot collaboration. For example, Wilcox et al. \cite{wilcox2013optimization} develop a scheduling paradigm that adapts to temporal disturbances and synchronizes activities. Pearce et al. \cite{pearce2018optimizing} develop an optimization for task assignment for a human-robot team that considers both time on task and ergonomic risk.

\subsection{End-User Programming}
End-user programming aims to reduce barriers for users without formal programming experience to participate in the technology development process. In robotics, end-user programming solutions have been proposed across a range of development phases; including during the setup, authoring, editing, and verification of robot programs \citep{ajaykumar2021survey}. The use of end-user programming in our work focuses on the authoring phase. Authoring solutions employ a variety of modalities to task the robot, including physical demonstrations \cite{paxton2017costar}, visual programming \cite{mateo2014hammer}, and extended reality. Our implementation uses screen-based augmented reality and focuses on iterative workflows where the operator provides either partial task specification or programming to complete complex tasks. Augmented reality has been leveraged in many prior authoring methods to provide contextualized visual programming \citep{cao2019v,gao2019pati,bambusek2019combining,luebbers2021arc}. For example, Akan et al. \cite{akan2011intuitive} use a robot-mounted camera to program manipulations (i.e., pick and place) of objects using a gripper-mounted robot camera. Several recent technologies focus on augmented reality interfaces through tablets or other mobile devices \citep{lambrecht2012spatial,frank2016realizing}. Our tangible interface was inspired by the Boston Dynamics Spot controller's touchscreen and game pad functionality \cite{Spot2023} and was focused on addressing common manufacturing challenges (e.g., registration, parameterization for surface-finishing tasks).

\subsection{Shared Autonomy}
Shared autonomy methods blend together the input of a human and robot policy to provide adjustable assistance \cite{selvaggio2021autonomy}. In this section, we mainly focus on examples of methods involving assistance based on goal inference, methods for dynamic role allocation, and methods to provide corrections to robot behaviors during execution. Dragan et al. \cite{dragan2013policy} proposed an early method where the system offered conditional assistance based on likely operator goals. Many works build on this formulation, such as to partially observable settings with unknown goals \cite{Javdani2018} and to completely remove ex-ante goals through a model-free deep reinforcement learning approach \cite{reddy2018shared}.

Dynamic role allocation methods use operator cues and task modeling to identify appropriate human interventions with a robot system. Medina et al. \cite{Medina2013} devise a dynamic assistance scheme based on measuring unexpected (i.e., disagreeing) interaction forces to inform between model-free and model-based assistance. Similarly, Evrard et al. \cite{Evrard2009} use physical interactions during collaborative tasks to shift agents between leader and follower roles using a homotopy-based controller. Li et al. \cite{Li2015} explore mixed role assistance by modeling the shifting assistance problem as a game theoretic two-agent system. Abbink et al. \cite{abbink2012haptic} explore practical factors related to role shifting, such as smooth transitions and hand-offs of assistance.

Finally, corrective methods aim to address variability in a task or environment. Many works rely on physical intervention to a robot manipulator to infer desired changes to a behavior or underlying policy \cite{nemec2018efficient,losey2017trajectory,bajcsy2017learning}. Most relevant to the system developed in this paper are methods that create intuitive real-time correction interfaces; including to modify the states of UAVs \citep{masone2014semi}, mobile robots \citep{cognetti2020perception}, and robotic manipulators \cite{hagenow2021corrective, hagenow2021informing}. The key difference from existing technology is that the corrective interface in this work is highly coupled with the uncertainty of task specification (e.g., proving a more expressive corrections interface when the robot behavior is coarsely defined through visual programming).
\section{Overview of Solution} \label{sec:prototype}
Industrial environments pose many challenges toward designing systems for human-robot teaming. To illustrate some of the key challenges, we consider a motivating example in aviation manufacturing before describing our system approach. 

\subsubsection{Motivating Example}
Imagine you are a factory worker on the final assembly floor for a wide-body aircraft. Many of your manual tasks, such as sanding, are highly variable. Each sanding task may have different amounts of material to be removed and may be subject to subtle variables including the sandpaper condition and wear of the sanding tool. Workers are incredibly adept at adjusting their work to react to these nuanced differences. Thus, an effective teaming solution should enable reactivity to the large degree of expected \emph{process variability}. Additionally, each task will have very different requirements. Some tasks may require following a predetermined procedure, whereas certain tasks, such as rework, may require working with engineers to plan a specific intervention. Thus, an effective teaming solution needs to enable \emph{flexible tasking}, where the end user can easily program or modify robot behaviors to match the specific needs of a task. Finally, over the course of a shift, you may spend time completing work in several areas of the aircraft. For example, you may perform surface prep for a large mating area of the fuselage sections and localized rework (e.g., sanding) across areas of the fuselage with excess composite resin. Thus, the platform requires \emph{mobility} to move with workers. 

\subsubsection{System Approach}
\begin{figure}[]
\centering
\includegraphics[width=0.48\textwidth]{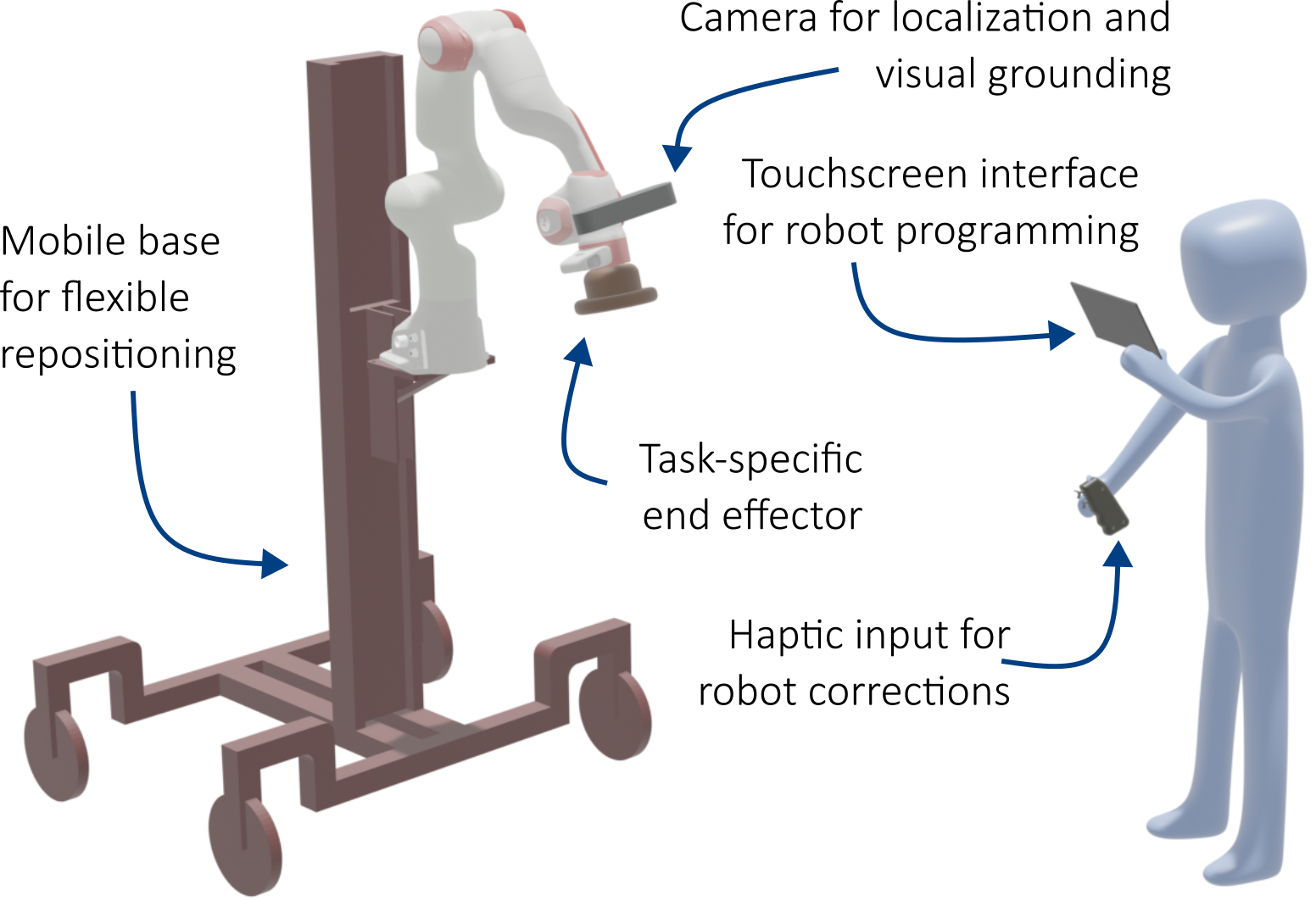}
\Description{A cartoon picture with the robot and operator that highlights key features of the mobile robot setup and the human-machine interface. The operator is holding a tablet and haptic input device.}
\caption{Proposed approach. The worker leverages a mobile augmented-reality interface and custom haptic device to program and augment a robot completing a task. The robot is on a mobile base and mounted with a RGB-D camera for localization and contextualized programming.}
\label{fig:capabilities}
\vspace{-15pt}
\end{figure}

To meet these requirements, we developed a system that tightly couples a skilled operator to an adjustable robot platform, as illustrated in Figure \ref{fig:capabilities}. The setup consists of a robot manipulator mounted to a mobile base. In this work, we consider a basic setup with casters that can be manually repositioned, however, future technologies could enhance mobility by mounting the manipulator on an automated ground vehicle \cite{kim2020moca}. The robot is equipped with a camera near the end effector to provide visual grounding during robot programming and localization between the robot platform and environment as the platform is relocated.

\textbf{End-user programming --} 
The robot is programmed by the worker through a mobile interface (e.g., tablet) where workers graphically specify tasks on an overlaid view from the robot camera \cite{senft2021task}. The operator's role is to select and register objects (in the case that the task model is known) and to select boundaries and parameters for the robot behavior when the task is unknown.

\textbf{Shared Autonomy --} 
We follow the approach of Hagenow et al. \cite{hagenow2021corrective} and allow differential robot corrections of the form:
\begin{align}
        \textbf{x} = \textbf{x}_{n} + \delta \textbf{x},\textbf{x}_{n} \in \mathbb{R}^{m}, \delta \textbf{y} \in S(\mathbb{R}^{m})
\label{eq:arbitration}
\end{align}
where $\textbf{x}$ is the final robot command (consisting of $m$ controlled variables), $\textbf{x}_{n}$ is a nominal command from a task model or user parameterization, and $\delta \textbf{x}$ is a differential correction applied by the operator. During execution, the role of the operator is to monitor the robot execution and intervene with the differential corrections (e.g., force, pitch, alignment, reversing) when the robot behavior performs poorly. The specifics of the corrections (including dimensionality and state variables) are discussed in the next section.

\section{Integration and Workflows} \label{sec:casestudy}
To better understand the potential of the proposed human-robot teaming approach, we developed a prototype implementation. While such a system should support a range of industrial tasks, we focused our initial implementation on sanding. The choice was mainly based on the prevalence of sanding tasks in industrial applications and the recent interest in flexible robot sanding platforms in the research literature \cite{kabir2018identifying,maric2020collaborative} and industry (e.g., GrayMatter Robotics \cite{Graymatter2023} and Norbo Robotics \cite{Nordbo2023}). Additionally, sanding tasks are a desirable candidate for human-robot teaming given the physical challenges of manual sanding and the high degree of process variability that makes broad robotic automation challenging.

\subsection{Tasks}
Our prototype sanding platform was designed to support two use cases that are based on the needs of common manual sanding tasks in aviation manufacturing. The first is the sanding of interior composite structures, such as section dividers and overhead bins. There is a large number of different geometries that furnish the fuselage interior. There is also a range of volumes from unique structures (e.g., forward galley) to higher volume, common geometries (e.g. window panels, overhead bins). Ideally, there exists enough repetition to partially automate such tasks. The geometries of the pieces are known a priori and while there are difficult curvatures (e.g., tight radii, concave features) and defects from upstream manufacturing processes, the availability of the task geometry and possibility to collect task data enables higher degrees of task planning.

The second task is inspired by fuselage rework. While the process of fiber layup has been mostly automated for composite fuselage sections, the composite structures still often require manual sanding following the curing process. After exiting the autoclave, fuselage sections are typically inspected for any problematic areas, for example if there are areas with excess resin buildup. The areas that require sanding will vary on a section by section basis and assume various shapes and sizes. Given that each sanding job is unique, this rework is less amenable to automation.

\begin{figure}[b]
\centering
\includegraphics[width=0.48\textwidth]{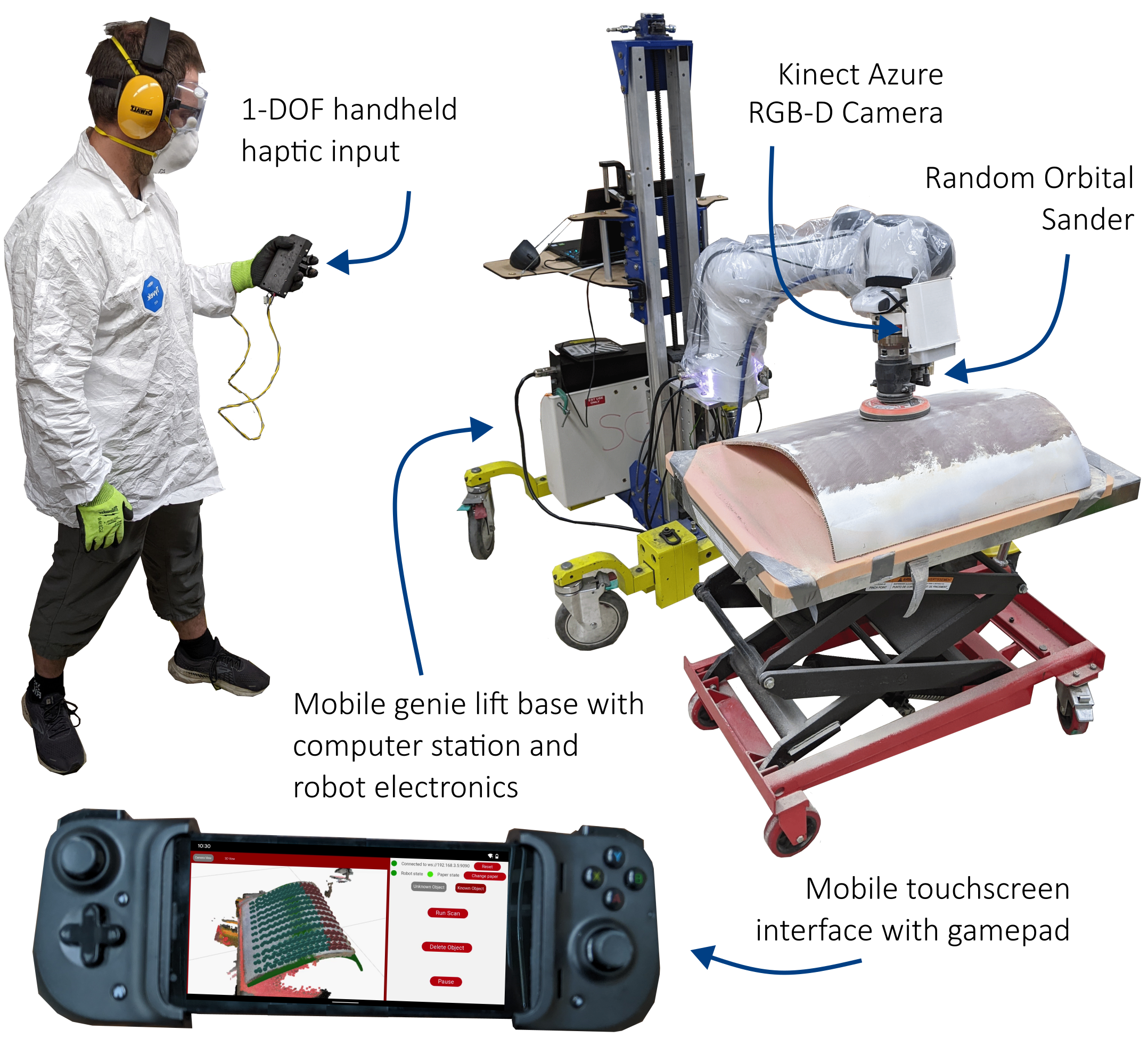}
\Description{A picture of the sanding robot, operator, and mobile interface with annotations for key features. The operator is holding the haptic device. The mobile interface has a game pad and shows a picture of the task piece.}
\caption{Overview of implementation. The worker's interface uses a touchscreen gamepad and (optional) low degree-of-freedom haptic device to interact with the robot. The robot has a custom stand and end effector for sanding.}
\label{fig:prototype}
\vspace{-15pt}
\end{figure}

\subsection{System Setup}
Our prototype setup is shown in Figure \ref{fig:prototype} and consists of a Franka Emika robot mounted to a modified genie lift. The lift allows for a worker to manually position the robot for a given task and make manual adjustments to the robot height through a jackscrew attached to the base of the robot. The robot is equipped with an ATI Axia80 force-torque sensor and a random orbital sanding tool as the end effector. The random orbital sander interfaces to the force-torque sensor via a 3d-printed flexible vibration isolator that helps to reduce the transmission of vibrations to the robot base. To facilitate better contact control with the environment, the robot is operated in a compliance control mode using the measured forces from the force-torque sensor. The orbital sander is powered pneumatically and toggled through a computer-controlled solenoid valve. Additionally, a Kinect Azure RGB-D camera on the robot's distal link is used for visual grounding and localization.

\begin{figure}[]
\centering
\includegraphics[width=0.48\textwidth]{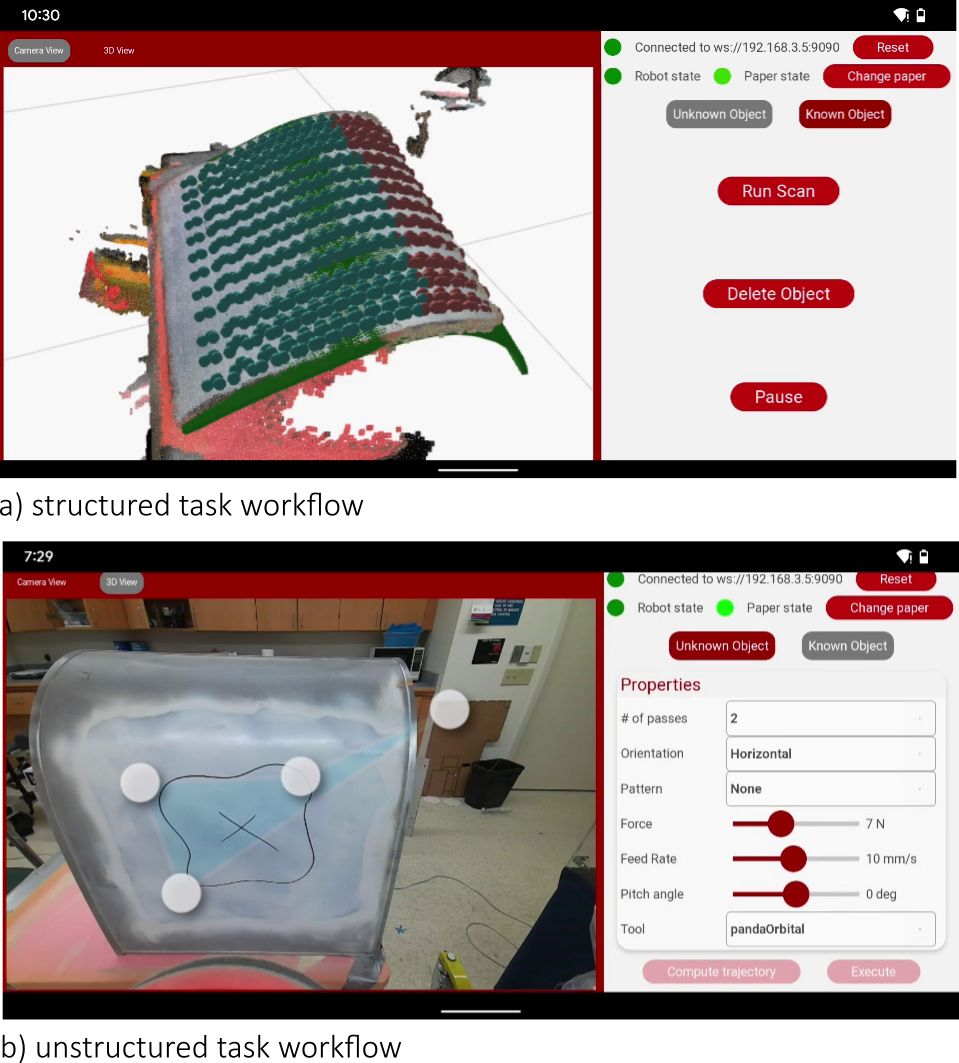}
\Description{In each picture, the interface has a left and right pane. For the structured workflow, the left pane shows the task geometry in 3D with the trajectory visualized. The right pane has some buttons to issue commands and indicators about the robot state. For the unstructured workflow, we see the task piece with four markers that highlight the area to be sanded. The right side shows sliders for selecting the robot parameters as well as similar indicators.}
\caption{Example interfaces for the structured and unstructured workflow. The structured workflow mostly leverages the scanned 3D view of the geometry. The unstructured workflow is programmed as an overlay on the current robot camera view. Both workflows include state indicators (e.g., robot state) and sandpaper monitoring.}
\label{fig:interfaceexamples}
\vspace{-15pt}
\end{figure}

The operator programs and interacts with the robot through a set of mobile interfaces. \footnote{Open source code (and CAD files) available at: \url{https://github.com/mhagenow01/panda_uli_demo}} The robot programming is achieved through a mobile touchscreen interface. The interface was developed using React javascript, and is served to the user via a webpage on a touchscreen mobile phone (Google Pixel 3a XL). The phone has an attached external gamepad to capture additional user input. 
The operator provides corrections either through the gamepad or optionally through a one degree-of-freedom haptic input \cite{doshi2023}.

 \begin{figure*}[]
\centering
\includegraphics[width=\textwidth]{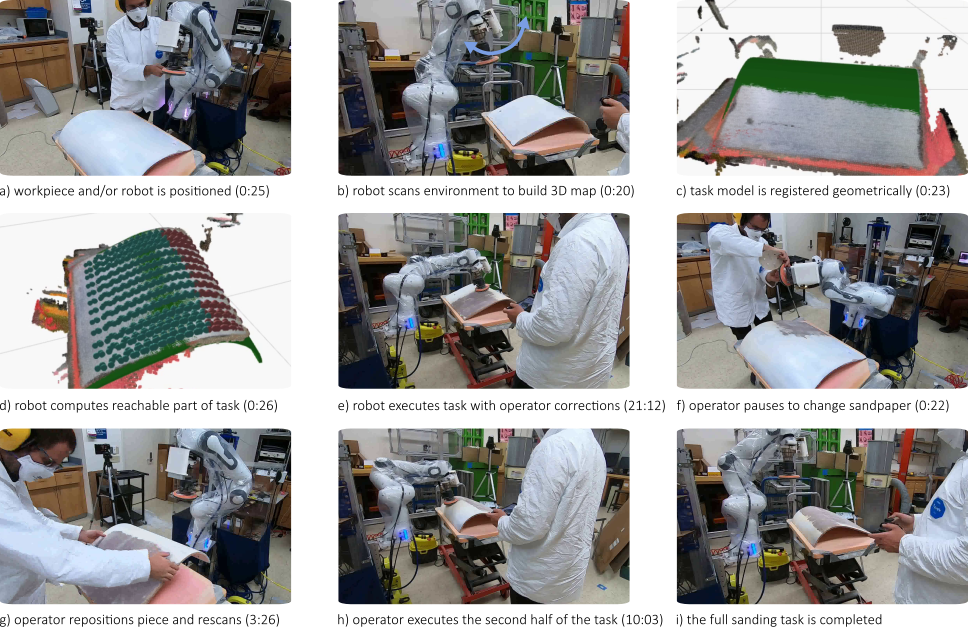}
\Description{The image shows nine pictures that highlight the structured task workflow and the captions report how long each step took. a) the workpiece and/or robot is positioned (25 seconds). b) the robot scans the environment to build a 3D map (20 seconds). c) the task model is registered geometrically based on the 3D scan (23 seconds). d) The robot compute and visualizes the reachable part of the task in 3D (26 seconds). e) the operator stands next to the robot while it executes and provides corrections through the mobile interface (21 minutes and 12 seconds). f) the operator pauses the robot to change the sandpaper (22 seconds). g) the operator repositions the workpiece by rotating it and rescans using the robot camera (3 minutes and 26 seconds). h). the person stands next to the robot and executes the second half of the task (10 minutes and 3 seconds). i) the person stands near the robot and can see the piece is fully sanded.}
\caption{Workflow for structured tasks. The task is to sand white paint off a curved composite structure. The workpiece is large enough such that it requires multiple configurations of sanding to complete. Event timing (MM:SS) is reported in parentheses.}
\label{fig:structuredworkflow}
\end{figure*}

\subsection{Supported Workflows}
The prototype system was designed to support two workflows inspired by the use cases in aviation manufacturing. The first use case is designed for more structured tasks, such as sanding the interior composite pieces, where it is possible to plan a priori and leverage previous task data to minimize the required input from the operator. The second use case is designed for less structured tasks, such as the composite fuselage rework, where few details of the task are known a priori and instead the focus is on allowing end-users to effectively specify the parameters for the desired sanding task. Both workflows use a common back-end which supports features such as reachability checking of the robot behavior poses, real-time corrections, and basic sandpaper monitoring (i.e., keeping track of the time the current sandpaper has been in use). Examples of the mobile interface for each workflow are shown in Figure \ref{fig:interfaceexamples}. The two workflows are explained in the following subsections, illustrated in Figures \ref{fig:structuredworkflow}-\ref{fig:unstructuredworkflow}, and shown in the supplementary video.

\subsubsection{Structured Tasks}
The high-level approach for structured sanding tasks (e.g., interior composite structures) uses a data-driven task model to identify the robot behavior and to inform the corrections an operator can make during the robot execution. The workflow assumes that the object geometry, task model (i.e., robot trajectories), and corrections that the operator can make are known.

The role of the operator is to position the robot and workpiece, register the task geometry, and correct the robot as the task model progresses. Given that not all areas of the workpiece may be reachable from a single static robot and workpiece configuration, the robot sands the reachable section of the task from its current configuration and keeps memory of the parts of the task that have been completed. While in some cases, it may be possible to complete an entire sanding task from one robot configuration, in many cases the geometry is larger than the dexterous robot workspace and thus, our workflow for structured tasks is iterative in nature, where the human re-positions the robot or workpiece (whichever is more feasible for a given task) multiple times to complete the overall task. The full workflow consists of the following steps:

\begin{enumerate}[leftmargin=*]
    \item The operator sets up the task by specifying the geometry (i.e., CAD model) and providing demonstrations with an instrumented tool (our implementation task model was hard-coded for simplicity) that are mapped to the object coordinate frame. 
    \item The robot and mobile base are moved into position for the sanding task. When the robot e-stop is engaged, the manipulator can be guided to a pose where the robot's end-effector camera can view the task geometry. The robot e-stop is disengaged which begins active control of the robot.
    \item Using the mobile interface, the operator presses a button to scan the environment. The robot performs panning motions relative to its current pose from which a 3D map of the environment is built  based on the robot-localized depth images \cite{labbe2019rtab}.
    \item The worker is presented with the 3D scan of the environment to identify the task geometry. The interface leverages a human-in-the-loop method for geometrically registering objects \cite{hagenow2022registering}. The system attempts to automatically determine the task model and object pose from the 3D scan. The operator verifies the fit and provides any required modifications, such as switching the geometry if it is incorrectly identified or adjusting the pose of the fit using the interface gamepad controls. Leveraging this registration process circumvents the need for custom tooling to fixture the workpieces in a known location, enabling more agile deployment to new tasks with varied geometries.
    \item Once the operator indicates a satisfactory fit, the system computes the portion of the robot task model that can be completed from the current workpiece and robot configuration (based on reachability). The robot path is overlaid on the task geometry with colors highlighting previously completed parts of the task (gray), reachable sanding paths (blue), and paths that cannot be reached from the current configuration (red). If the operator is satisfied, they can proceed with the sanding. If not, the operator can reposition the workpiece and restart the workflow.
    \item The robot executes the sanding task. As the robot executes, the operator provides input. The operator can interrupt the workflow if the sanding disc is too worn, which pauses the sanding, moves the robot to a configuration where the sanding disc can easily be changed, then resumes the workflow where it left off. The operator can issue corrections to the robot as it sands. Given that the robot task model is based on previous successful task data, we aim to decrease the need for and complexity of operator corrections. For example, in our implementation the operator can provide corrections to the abrasiveness when sanding areas with defects (a combination of speed, force, and tool pitch). The operator can also reverse the execution (i.e., backtrack) using a button when the sanding is insufficient.
    \item Once the sanding is complete, the interface changes the color of the completed portion of the task from blue to gray. If there are still remaining areas to be sanded, the operator can then reposition the workpiece (e.g., rotating the piece 180 degrees to sand the other side) and iterate through the workflow steps.
\end{enumerate}

\subsubsection{Unstructured Tasks}
The approach for unstructured tasks (e.g., sanding a defect on the fuselage) combines a graphical method for task specification with real-time user corrections. The premise is to further leverage the expertise of the operator to compensate for the lack of prior knowledge about the requirements of the sanding. In this way, the unstructured task workflow can be more easily applied to new sanding tasks, but requires more operator effort. The role of the operator is to position the robot and workpiece, identify the bounds of the required sanding and associated sanding parameters, and to provide corrections as the robot executes the task. The full workflow consists of the following steps:

\begin{enumerate}[leftmargin=*]
    \item Similar to the first workflow, the robot camera is positioned to provide a satisfactory view of the desired sanding area.
    \item The worker directly programs the desired sanding on the mobile interface. The worker annotates the sanding bounds by positioning a set of markers on top of the robot camera view. As the target sanding area is moved, the system provides reachability checks on the interface as a grid of points that are colored based on whether the point is reachable (green) or not (red).
    \item The operator specifies desired sanding parameters, such as the number of passes, orientation (i.e., horizontal or vertical), applied force, tangential velocity, and tool pitch.
    \item Once the operator is satisfied with the sanding parameterization, the execution is started and the operator provides feedback. Similar to the structured workflow, the operator can change sandpaper when necessary and issue real-time corrections to the robot behavior. Because the task model is programmed and not data-driven, the operator may need to provide corrections to any robot variable which is accomplished by mapping the gamepad joysticks and triggers to corrections to different robot state variables (e.g., force, path, pitch).
    \item If the task is completed successfully, there is no additional work for the operator. If the sanding was insufficient, the operator can trivially execute the same sanding program a second time. The operator can also make any required modifications, such as adjusting the sanding area or parameters. Thus, this workflow can also be completed iteratively for a given sanding task.
    
\end{enumerate}

 \begin{figure*}[]
\centering
\includegraphics[width=\textwidth]{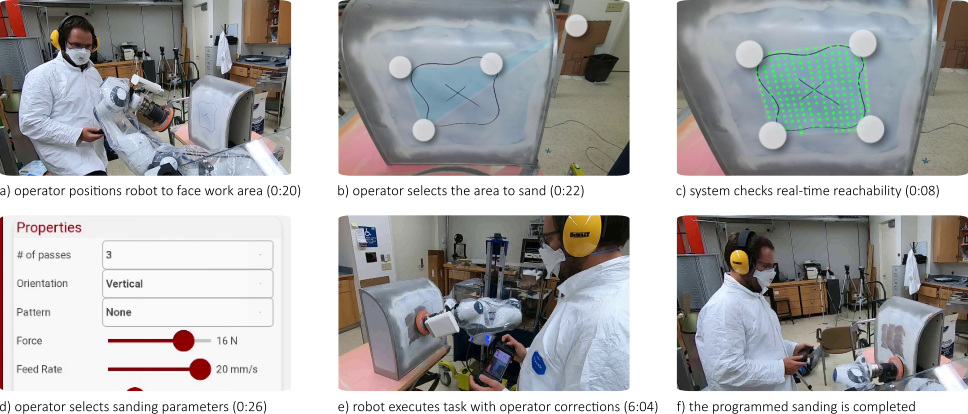}
\Description{The image shows six pictures that highlight the unstructured task workflow and the captions report how long each step took. a) the operator moves the end effector of the robot so the camera faces toward the upright piece that needs to be sanded (20 seconds). b) the operator uses four circular markers on a touchscreen to identify the area that the robot needs to sand (22 seconds). c) the system shows a green grid of points within the selected area that indicate that each point is reachable by the robot (8 seconds). d). close up of the sliders and drop downs the operator uses to select sanding parameters including number of passes, orientation, pattern, force, and feed rate (26 seconds). e) the operator stands behind the robot while it sands and gives corrections through the mobile input (6 minutes and 4 seconds). f) the operator stands near the piece and confirms it is fully sanded.}
\caption{Workflow for unstructured tasks. The task is to sand the white paint in the area highlighted in sharpie marker. Event timing (MM:SS) is reported in parentheses.}
\label{fig:unstructuredworkflow}
\end{figure*}

\subsection{Case studies}
Representative tasks were constructed for both workflows to demonstrate the proposed prototype system (as seen in Figures \ref{fig:structuredworkflow}-\ref{fig:unstructuredworkflow}). The system was tested in two ways. First, we conducted a series of informal lab tests (as shown in the supplementary video). Second, we arranged a series of on-site sessions where the proposed system and workflows were demonstrated to end users and engineers in an aviation manufacturing facility. Due to confidentiality agreements, we are unable to report specific data from these on-site demonstrations and instead generalize the findings in the next section.

For the structured worfklow, the task consisted of sanding spray paint off of a curved composite structure. The structure was large enough such that the robot could not sand the entire piece from a single configuration and thus the task requires two iterations of the workflow. While the robot behavior and corrections could be learned directly from expert demonstrations \cite{hagenow2021informing}, for simplicity, the robot task behavior used a hard-coded set of passes over the surface. The corrections the operator could make were limited to be one-dimensional (a combination of force, speed, and pitch) such that they could be provided using a one-degree-of-freedom input.

For the unstructured workflow, the task consisted of removing spray paint from a vertical composite structure (a similar orientation to what would be expected during fuselage sanding) with an identified area for sanding. The area to be sanded was marked using a sharpie marker similar similar to the inspection process for composite rework. Given that the robot behaviors were specified using a basic parameterization, the operator was able to provide corrections to all key robot state variables (e.g., force, path, speed) independently using the joysticks and triggers on the mobile interface gamepad. While this increased control space complicates the required user input during robot execution, the required expressiveness of the corrective input is a consequence of the lack of task knowledge and simplicity of the programming interface.
\section{Discussion} \label{sec:discussion}
In this section, we summarize the results from assessing our system in the lab setting. We then discuss the outcomes and lessons learned related to shared autonomy and end-user programming in our prototype system. Finally, we describe key end-user feedback from our on-site session and resulting opportunities identified for the development of future human-robot teaming systems.

We generally found that the proposed system achieved appropriate sanding quality (i.e., visibly removing all paint in the expected regions) for the case study tasks in the lab setting. In a small few instances, there were regions that were missed (e.g., boundaries between sanding iterations or edges), however, we envision addressing these challenges by (1) tuning the overlap of broken-up behaviors or (2) accepting there will be some manual cleanup required as, in practice, perfect automation would be too costly to achieve. No concerns about sanding quality were raised during our on-site demonstrations, which included the two case study tasks and two additional surface preparation tasks.

Both EUP and shared autonomy were critical technologies in the performance of the prototype system. EUP was required by design for successful use of our system (i.e., each workflow had required steps for the user to program at the task level). In the structured workflow, we found that operator input through EUP was crucial during registration (when the object was geometrically non-unique) and in iterating on workpiece placement based on reachability. For the unstructured workflow, the visual programming enabled fast, iterative specification of robot behaviors. Given the coarse specification (i.e., selecting the boundary on the touch screen), such a technique would be best suited for tasks without needs for a precision boundary (i.e., rework where a margin of sanding around the damage is desired). We also observed that real-time operator corrections were critical to achieve quality sanding, even for a well tuned task model. For the structured workflow, the task model was sufficient to achieve coverage and operator corrections could focus on the sanding quality in each area. Even with a tuned task model, we found that corrections were needed for the majority of the task execution. These included corrections to the pitch to modulate the material removal rate and to reverse and repeat the execution for areas with excess paint. Without corrections, the robot would undersand and oversand some areas. For the unstructured task, we found found similar value in high-frequency corrections, such as repeating passes with updated paths and adjusting the tool pitch.


Our feedback from industrial technologists during the on-site demonstration identified three key limitations for future study.
\begin{enumerate}[leftmargin=*]
\item \textbf{The collaborative robot solution is too slow and small. Other industrial sanding solutions can sand wider areas much faster for surface-scuffing (i.e., light sanding) tasks.}
Collaborative robots are designed to work safely alongside human workers. However, this safety comes at the cost of robot size and speed. As discussed in the introduction of the prototype platform, the dexterous reach of collaborative robot platforms is limited and often prohibitive for surface-finishing tasks. In particular, if the duration of the work is short (e.g., scuff sanding), the preliminary process of setting up and resetting the robot greatly affects efficiency. Additionally, for tasks involving large, fast motions, allocating the work to a collaborative robot (e.g., breaking up the task, slower completion times) might be inefficient. We believe it is crucial for adoption to develop a model to assess whether a manufacturing task is well suited to the cobot's capabilities and limitations. For example, a composite scarf repair may be amenable to a collaborative robot solution as the sanding task typically involves localized, low-force, and prolonged sanding to penetrate the many composite layers.

\item \textbf{It would be useful if the technology business case included both improved ergonomics and improved worker efficiency.} Our solution clearly demonstrated improved physical ergonomics by distancing the worker from the force and vibration loads associated with manual sanding during high variability tasks. However, compared to existing sequestered robot sanding solutions where workers can set up tasks and then complete secondary responsibilities, our workflows required significantly greater user involvement (i.e., time). Coupled with decreases in speed induced by the the platform velocity limits, there was a desire to investigate ways to improve worker efficiency. We imagine achieving this increased efficiency through several means. First, we believe it may be possible to increase the automation of the unstructured workflow by employing automated suggestions for parameterization. For example, if the damage is outlined or we could build a database of previous sanding instances, it might be possible to classify the likely sanding parameters and only require human input for verification and corrections (similar to the registration process). Additionally, for tasks requiring intermittent corrections (i.e., difficult curvatures of sanding), we envision scaling opportunities where a worker supervises multiple robots \cite{hagenow2023coordinated}.

\item \textbf{Future research should focus on the interaction mechanisms and workflows, rather than the full system development.}
We found that the graphical interface and reduced-dimensionality corrective input garnered the most interest during the on-site demonstration. While an original system goal was to design a flexible platform for sanding tasks, specifically for tasks that could leverage the same type of end-of-arm tooling (e.g., a random orbital sander), we found that users were less interested in the platform flexibility (i.e., completing multiple types of work as needed by factory workers) due to a number of environmental and certification challenges that limit the broad applicability of a human-robot teaming solution. For example, a painting application requires different hardware (e.g., electrostatic protection) from a setup 
 for working in confined spaces (e.g., kinematics and payload). Thus, rather than focusing on the full robot platform , we believe future work should focus on a human-machine interface for flexible tasking with flexible hardware (e.g., manipulators tailored to specific factory tasks). Development and adoption of such an interface could potentially decrease required training for workers completing many different jobs across the factory. For example, we imagine using the same tools developed in this work to enable robotic fastener insertion, where the operator uses end-user programming to select locations for fastener installations and provides low-level corrections to address alignment error during the insertions.

\end{enumerate}

\subsection{Limitations \& Future Work}
In this section, we discuss the limitations of our human-robot teaming system. Our approach was only evaluated informally. Going forward, user studies are needed to quantify the benefits of the proposed approach and workflows. This includes studies that measure the impact of the human-robot teaming solution on performance and ergonomics as well as studies that estimate end-user acceptance \cite{davis1989perceived} by evaluating the system with representatives of the target user populations, including users of varied expertise. Additionally, building toward our vision of flexible human-robot teaming will require evaluating our approach across a range of physically demanding and variable tasks (i.e., evaluations beyond sanding -- such as fastening/assembly and composite layup). In addition to tasks where the robot assumes full execution, the approach can be extended to consider tasks that require interdependent tasks by the robot and the worker (e.g., highly dexterous and low risk sanding where the operator can outperform the robot) \cite{senft2021situated}.

There are also limitations in our implementation that we plan to address in future system iterations. First, for assessment in realistic industrial applications, significant efforts are needed to raise the technological readiness level of the prototype system (e.g., computational efficiency and robustness). Second, each workflow contained tools related to reachability that were solved point-wise (i.e., not considering path-wise continuity). Future work will explore how to include better tools for reachability that balance accuracy with speed for use in human-in-the-loop workflows. Regarding mobility, our case study focused on tasks where the workpiece was repositioned (e.g., interior structures), but did not investigate repositioning the robot platform (e.g., large fuselage sections). Similarly, our approach to monitor sandpaper health was based on a simple model and not formally validated. In the future, we would like to explore predictive methods that may enable better sanding through accurate sandpaper tracking. Finally, the robot behavior for our structured workflow was manually specified. Future work should explore how end users can provide demonstrations to the robot.

  
\section{Conclusion}
In this work, we proposed a system for human-robot teaming that leverages end-user programming and shared autonomy and implemented an instantiation of our approach for sanding tasks. In our approach, the operator is engaged throughout the full task workflow, including the initial programming and specification of the task as well as during the robot's execution. The implementation, contextualized in sanding, involved two workflows targeted toward more and less structured tasks where the operator interacts with the robot via an augmented-reality tablet interface and custom haptic device. We designed representative lab tasks to demonstrate each workflow and discussed takeaways from our testing with aviation manufacturing stakeholders, including remaining challenges and recommendations for driving human-robot teaming forward in physically demanding industrial applications.

\begin{acks}
This work was supported by the Grainger Wisconsin Distinguished Graduate Fellowship and a NASA University Leadership Initiative (ULI) grant awarded to the UW-Madison and The Boeing Company (Cooperative Agreement \# 80NSSC19M0124). We also thank the Wisconsin Alumni Research Foundation for the initial system photo.
\end{acks}


\bibliographystyle{ACM-Reference-Format}
\balance
\bibliography{references}



\end{document}